\documentclass[10pt,twocolumn,letterpaper]{article}

\usepackage[pagenumbers]{cvpr} 

\usepackage{graphicx}
\usepackage{amsmath}
\usepackage{amssymb}
\usepackage{booktabs}
\usepackage{newfloat}
\usepackage{listings}
\usepackage{colortbl}
\usepackage{multirow}
\usepackage{algorithm}
\usepackage{algorithmic}
\usepackage{graphicx}
\usepackage{booktabs}
\usepackage{multirow} 
\usepackage{overpic}
\usepackage{hhline}
\usepackage{arydshln} 
\usepackage{pifont}
\usepackage{enumitem}
\usepackage{chapterbib}
\usepackage{colortbl}
\usepackage{makecell}
\usepackage{xspace}

\definecolor{cvprblue}{rgb}{0.21,0.49,0.74}
\usepackage[pagebackref,breaklinks,colorlinks,allcolors=cvprblue]{hyperref}

\def\paperID{*****} 
\def\confName{CVPR}
\def\confYear{2025}

\title{QMambaBSR: Burst Image Super-Resolution with Query State Space Model}

\author{
Xin Di\textsuperscript{1$\dagger$}, 
Long Peng\textsuperscript{1$\dagger,\ddagger$}, 
Peizhe Xia\textsuperscript{1}, 
Wenbo Li\textsuperscript{2}, 
Renjing Pei\textsuperscript{2}\thanks{Renjing Pei and Yang Wang are the corresponding authors. $\dagger$ These authors contributed equally to this work. $\ddagger$Long Peng is the project leader. This work was finished during Xin Di and Long Peng in the internship of Huawei Noah.}, 
Yang Cao\textsuperscript{1}, 
Yang Wang\textsuperscript{1,3*}, 
Zheng-Jun Zha\textsuperscript{1} \\
\textsuperscript{1}University of Science and Technology of China, 
\textsuperscript{2}Huawei Noah's Ark Lab, 
\textsuperscript{3}Chang'an University \\
{\tt\small \{dx9826, longp2001\}@mail.ustc.edu.cn},  
{\tt\small peirenjing@huawei.com}, 
{\tt\small ywang120@chd.edu.cn}
}

\begin{document}
\maketitle
\begin{abstract}
Burst super-resolution (BurstSR) aims to reconstruct high-resolution images by fusing subpixel details from multiple low-resolution burst frames. The primary challenge lies in effectively extracting useful information while mitigating the impact of high-frequency noise. Most existing methods rely on frame-by-frame fusion, which often struggles to distinguish informative subpixels from noise, leading to suboptimal performance. To address these limitations, we introduce a novel Query Mamba Burst Super-Resolution (QMambaBSR) network. Specifically, we observe that sub-pixels have consistent spatial distribution while noise appears randomly. Considering the entire burst sequence during fusion allows for more reliable extraction of consistent subpixels and better suppression of noise outliers. Based on this, a Query State Space Model (QSSM) is proposed for both inter-frame querying and intra-frame scanning, enabling a more efficient fusion of useful subpixels. Additionally, to overcome the limitations of static upsampling methods that often result in over-smoothing, we propose an Adaptive Upsampling (AdaUp) module that dynamically adjusts the upsampling kernel to suit the characteristics of different burst scenes, achieving superior detail reconstruction. Extensive experiments on four benchmark datasets—spanning both synthetic and real-world images—demonstrate that QMambaBSR outperforms existing state-of-the-art methods.
\end{abstract}

\section{Introduction}

{In recent years, with the continuous development of smartphones, overcoming the limitations of smartphone sensors and lenses to reconstruct high-quality, high-resolution (HR) images has become a research hotspot. Benefited from the development of deep learning, single image super-resolution (SISR)~\cite{ju2023mean,a:5,a:6,a:7,peng2024efficient,peng2024towards} has achieved remarkable progress, but the performance is still limited by the finite information provided by a single image. Consequently, numerous researchers are dedicating their efforts to burst super-resolution (BurstSR), which aims to leverage the rich sub-pixel details provided by a sequence of burst low-resolution images captured by hand-tremor and camera/object motions to overcome the limitations of SISR, achieving substantial advancements~\cite{a:8,a:9,a:10,a:21,a:22}.}

\begin{figure}[!t]
	\centering
	\includegraphics[width=1.0\linewidth]{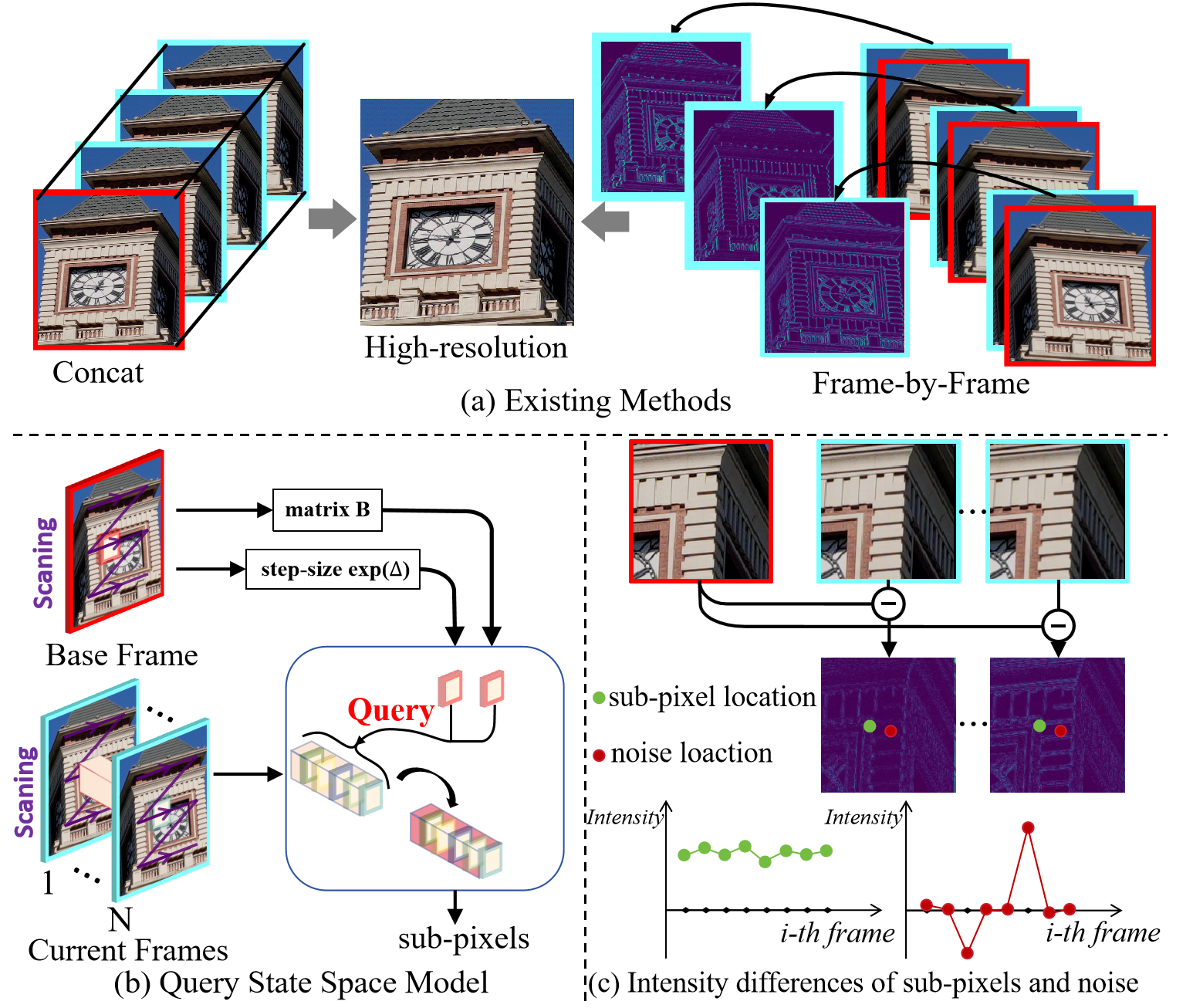}

\caption{The concat and frame-by-frame operations in existing methods struggle to efficiently extract sub-pixels and suppress noise, leading to remaining artifacts and over-smoothed details, as shown in (a). We observe that noise randomly appears on several frames, while effective sub-pixels have consistent intensity at corresponding positions in all frames, as shown in (c). Based on this, a novel inter-frame query and intra-frame scanning-based QMambaBSR is proposed to extract more accurate sub-pixels while mitigating noise interference simultaneously, as shown in (b).}
	\label{fig:intro}
\end{figure}

In BurstSR, the first image is the super-resolution frame, denoted as the base frame, while the remaining images, referred to as current frames, supply sub-pixel information for producing a high-quality HR image. The pipeline of most existing BurstSR approaches can be mainly categorized: alignment, fusion, and upsampling. Firstly, due to the misalignment caused by hand-tremor, alignment methods~\cite{bhat2021deep,wei2023towards,c:6,a:24} are employed to align the current frames with the target base frame. Then, the primary challenge lies in extracting sub-pixel information from the current frames that match the content of the base frame while concurrently suppressing high-frequency random noise. Previous methods, such as weighted-based multi-frame fusion~\cite{bhat2021deep,a:23}, obtain residuals by subtracting each current frame from the base frame and utilizing simple weighting techniques to fuse obtained residual information. Although easy to perform, these methods neglect the inter-frame relationship among multi-frames, failing to extract sub-pixels that better match the base frame and are susceptible to interference from noise in RAW images. To enhance inter-frame relationships, BIPNet~\cite{c:6} proposes channel shuffling of multi-frame features to improve information flow between different frames. Consequently, recent state-of-the-art methods, such as GMTNet ~\cite{dudhane2023burstormer,c:7,luo2021ebsr}, propose using cross-attention, explicitly employing the base frame as a query to retrieve and capture feature differences from the current frame pixel-to-pixel to extract sub-pixel information. RBSR ~\cite{wu2023rbsr} utilize RNNs~\cite{liang2015recurrent} for frame-by-frame feature fusion, as shown in Figure~\ref{fig:intro} (a). The aforementioned methods extract sub-pixel information and denoise in a frame-by-frame manner or implement sub-pixel extraction and denoising through separate modules. However, these methods overlook the distinct characteristics between sub-pixels and noise, making it challenging to capture useful sub-pixels from noise and leading to artifacts and over-smoothed details.

After fusion, adaptively learning high-resolution mappings from the extracted and fused features remains a paramount challenge in BurstSR. The existing state-of-the-art methods, such as Burstormer ~\cite{dudhane2023burstormer} and BIPNet~\cite{c:6}, primarily utilize static interpolation, transposed convolution~\cite{gao2019pixel}, or pixel shuffle~\cite{shi2016real} for static upsampling. Nevertheless, these approaches make it difficult to adaptively perceive the variations in sub-pixel distribution across different scenes by employing static upsampling methods, resulting in the inability to utilize the spatial arrangement of sub-pixels to accurately reconstruct high-quality, high-resolution (HR) images.

To address these issues, we propose a novel Query Mamba Burst Super-Resolution (QMambaBSR) network, which integrates a novel Query State Space Model (QSSM) and an Adaptive Up-sampling module (AdaUp) to reconstruct high-quality high-resolution images from burst low-resolution images. Specifically, QSSM is first proposed to efficiently extract the sub-pixels in both inter-frame and intra-frame while mitigating noise interference. In particular, QSSM retrieves information across current frames for the base frame by modifying control matrix \(B\) and discretization step size $\Delta$ in the state space function, as shown in Figure~\ref{fig:intro} (b). QSSM synchronously performs inter-frame querying and intra-frame scanning, comprehensively considering both inter-frame and intra-frame information to achieve integrated sub-pixel extraction and noise suppression. AdaUp is proposed to perceive the spatial distribution of sub-pixel information and subsequently adaptively adjust the upsampling kernel to enhance the reconstruction of high-quality HR images across diverse burst LR scenarios. Furthermore, to comprehensively fuse sub-pixels with different scales, the Multi-scale Fusion Module is proposed to combine channel  Transformer and local CNN, as well as horizontal and vertical global Mamba, to fuse sub-pixel information of different scales. Extensive experiments on four popular synthetic and real-world benchmarks demonstrate that our method achieves a new state-of-the-art, delivering superior visual results.

The contributions can be summarized as follows:
\begin{itemize}
    \item {A novel inter-frame query and intra-frame scanning-based Query State Space Model (QSSM) is proposed to extract more accurate sub-pixels while mitigating noise interference simultaneously.}
    \item {We propose a novel Adaptive Up-sampling module, designed respectively for adaptive up-sampling based on the spatial arrangement of sub-pixel information in various burst LR scenarios, and for the fusion of sub-pixels across different scales.}
    \item {Our proposed method achieves new state-of-the-art (SOTA) performance on the four popular public synthetic and real benchmarks, demonstrating the superiority and practicability of our method.}
\end{itemize}

{\section{Related Work}
In this section, we briefly review Multi-Frame Super-Resolution and State Space Models. More comprehensive surveys are provided in ~\cite{xu2024survey,bhat2022ntire}.}

{\noindent\textbf{Multi-Frame Super-Resolution.} With the rapid development of deep learning in recent years~\cite{a:1,a:2,a:3,a:4}, deep-learning-based single image super-resolution (SISR) achieves significant breakthroughs~\cite{a:13,a:14,a:15,a:16}. However, due to the limited information provided by a single image, the performance of SISR is significantly constrained~\cite{a:11,a:12,a:17,a:18}. Therefore, Multi-Frame Super-Resolution (MFSR) is proposed to overcome the limitations of SISR by leveraging the useful sub-pixel information contained in multiple low-resolution images, achieving superior high-resolution reconstruction. In particular, DBSR~\cite{bhat2021deep} proposes using optical flow methods to explicitly align multiple low-resolution images and then fuse their features through attention weights. MFIR~\cite{a:23} utilizes optical flow for feature warping and proposes a deep reparametrization of the classical MAP formulation for multi-frame image restoration. BIPNet~\cite{c:6} proposes a pseudo-burst fusion strategy by fusing temporal features channel-by-channel, enabling frequent inter-frame interaction. Burstormer~\cite{dudhane2023burstormer} leverages multi-scale local and non-local features for alignment and employs neighborhood interaction for further inter-frame feature fusion. RBSR~\cite{wu2023rbsr} utilizes recurrent neural networks for progressive feature aggregation. However, most of these methods mainly use frame-by-frame approaches or pairwise interactions, either failing to explicitly extract sub-pixel information from the current frames or only querying the current frame point-by-point from the base frame. This makes it difficult for them to effectively extract sub-pixel details while suppressing noise interference. To address these limitations, we propose Query Mamba Burst Super-Resolution (QMambaBSR), which allows the base frame to simultaneously query inter-frame and intra-frame information to extract sub-pixel details embedded in structured regions while also suppressing noise interference.}

{\noindent\textbf{State Space Models.} State Space Models (SSMs) originated in the 1960s in control systems~\cite{a:25}, where they are used for modeling continuous signal input systems. Recently, advancements in SSMs have led to their application in computer vision~\cite{a:30,patro2024simba,fu2024ssumamba,chen2024changemamba}. Notably, Visual Mamba introduced a residual VSS module and developed four scanning directions for visual images, achieving superior performance compared to ViT~\cite{a:31} while maintaining lower model complexity, thereby attracting significant attention~\cite{c:8,a:32,a:33,a:34,a:35,a:36}. QueryMamba~\cite{zhong2024querymamba} is proposed to apply SSM to video action forecasting tasks. MambaIR~\cite{c:8} is the first to employ SSMs in image restoration, enhancing efficiency and global perceptual capabilities. However, there remains potential for further exploration of SSMs in BurstSR. Therefore, we propose a novel Query-based State Space Model designed to efficiently extract sub-pixel information for BurstSR.}

\section{Method}

\subsection{Overview}
Given a sequence of input low-resolution (LR) images, denoted as $\{x_i\}_{i=1}^N$, where $N$ represents the number of burst LR frames. Following~\cite{bhat2021deep}, we denote the first image as the base frame for super-resolution, while the other LR frames are used to provide rich sub-pixel information and are referred to as current frames. BurstSR can be defined as utilizing the sub-pixel information extracted from the current frames to supplement the base frame, generating a high-quality, high-resolution RGB image $I_{HR}$ with a super-resolution factor of $s$.

To achieve this goal, we propose QMambaBSR for burst image super-resolution, as illustrated in Figure~\ref{fig:framework}. First, we use alignment block~\cite{dudhane2023burstormer} to align the current images to the spatial position of the base frame. Next, we introduce a novel Query State Space Model (QSSM) designed to query sub-pixel information from the current images and mitigate noise interference in both inter-frame and intra-frame manner. Additionally, we present a novel Adaptive Up-sampling (AdaUp) module, which facilitates adaptive up-sampling based on the spatial arrangement of sub-pixel information in various burst images. Finally, a new Multi-scale Fusion Module is incorporated to fuse sub-pixel information across different scales. Next, we provide a detailed explanation of each component.

\subsection{Query State Space Model}
\begin{figure*}[!t]
	\centering
	\includegraphics[width=1.0\linewidth]{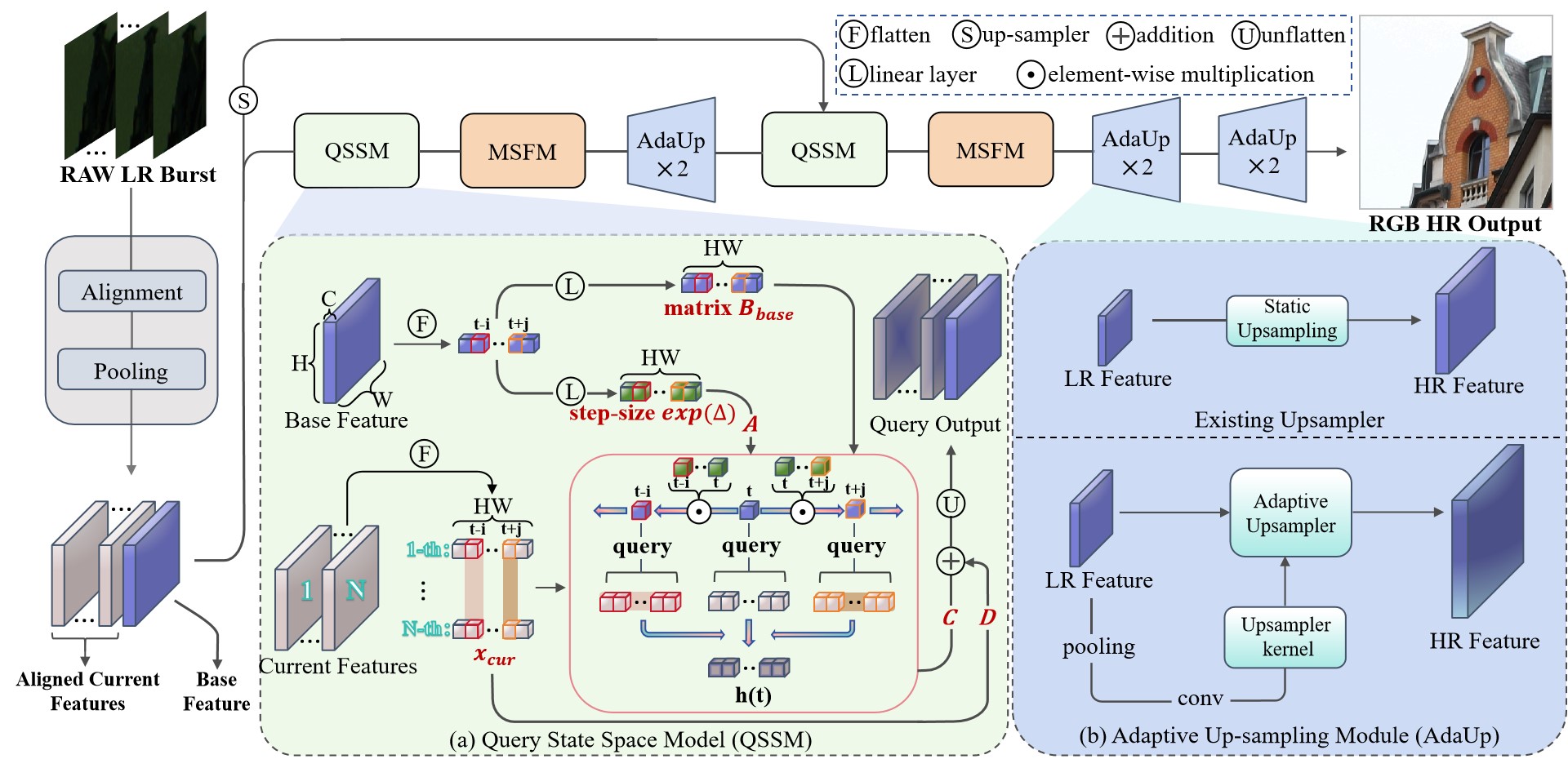}
	\caption{The overall framework of our proposed QMambaBSR, primarily includes the novel Query State Space Model (QSSM), Multi-scale Fusion Module (MSFM), and the Adaptive Up-sampling Module (AdaUp).}
	\label{fig:framework}
\end{figure*}
Considering that burst RAW images often contain high-frequency random noise and the sub-pixel information to be extracted typically shares a similar distribution with the base frame, it is crucial for the BurstSR task to utilize the base frame to uncover the rich sub-pixel information contained in the current frames for super-resolution, while simultaneously suppressing noise. Existing methods~\cite{bhat2021deep,dudhane2023burstormer,wei2023towards,wu2023rbsr,c:6} simply concatenate multi-frame information but fail to precisely extract sub-pixel information from the current frames using the base frame, resulting in a scarcity of sub-pixel details and consequently making it difficult to reconstruct fine details. Moreover, while some existing approaches attempt to use cross-attention~\cite{c:7}, utilizing the base frame as the query for sub-pixels extraction, such frame-by-frame methods struggle to suppress noise interference and are plagued by high computational complexity. This often results in the presence of noise and the introduction of artifacts. Therefore, we propose the Query State Space Model (QSSM), which enables the base frame to efficiently query all current frames simultaneously in both intra-frame and inter-frame manners. By leveraging the consistent distribution of sub-pixels and the inconsistent distribution of noise, our QSSM can simultaneously query multiple images to extract the necessary sub-pixel information while effectively suppressing random noise.

First, let's briefly review the State Space Model (SSM). The latest advances in structured state space sequence models (S4) are largely inspired by continuous linear time-invariant (LTI) systems, which map input \( x(t) \) to output \( y(t) \) through an implicit latent state \( h(t) \in \mathbb{R}^N \) ~\cite{c:8}. This system can be represented as a linear ordinary differential equation (ODE):
\begin{equation}
\begin{aligned}
    \dot{h}(t) &= Ah(t) + Bx(t), \\
    y(t) &= Ch(t) + Dx(t),
\end{aligned}
\label{eq:1}
\end{equation}
where \(N\) is the state size, \(A \in \mathbb{R}^{N \times N}\), \(B \in \mathbb{R}^{N \times 1}\), \(C \in \mathbb{R}^{1 \times N}\), and \(D \in \mathbb{R}\). Discretized using a zero-order hold as follows:
\begin{equation}
\begin{aligned}
    \overline{A} &= \exp(\Delta A), \\
    \overline{B} &= (\Delta A)^{-1} (\exp(\Delta A) - I) \Delta B,
\end{aligned}
\label{eq:2}
\end{equation}
After the discretization, the discretized version of Eq.~\eqref{eq:1} with step size \(\Delta\) can be rewritten as:
\begin{equation}
\begin{aligned}
    h_k &= \overline{A} h_{k-1} + \overline{B} x_k, \\
    y_k &= Ch_k + Dx_k,
\end{aligned}
\label{eq:3}
\end{equation}
At this point, the LTI system's input parameter matrices are static. Therefore, recent work~\cite{c:8}  makes \(B\), \(C\), and \(\Delta\) depend on the input. Recent research suggests that since the \(A\) matrix is a HIPPO matrix and \(\Delta\) represents the step size, \(\exp(\Delta A)\) can be viewed as forget gate and input gate~\cite{c:9}, which modulates the influence of the input on the state. 

However, traditional SSM lacks the multi-frame querying capabilities that are crucial for BurstSR tasks. Therefore, we propose a QSSM to enable the base frame to gate the output of the current frames, thereby allowing the base frame to perform information queries on the current frames to obtain the sub-pixels while eliminating noise, as illustrated in Figure~\ref{fig:intro}. Specifically, we let the current frames drive the state changes, with the base frame gating the influence of the current frames on the state through \(B\) and \(\Delta\). As shown in Figure~\ref{fig:framework} (a), the corresponding formulas are as follows:
\begin{equation}
\begin{aligned}
  h_t &= (\overline{A}_{\text{base}_t}) h_{t-1} + (\overline{B}_{\text{base}_t}) x_{\text{cur}_t}, \\
  y_t &= Ch_t + Dx_{\text{cur}_t}, 
\end{aligned}
\label{eq:4}
\end{equation}

Considering that noise may exist at certain locations in the base frame and is randomly distributed at the same locations in other current frames, we concatenate the base frame with other current frames to initially reduce random noise in the original base frame and achieve preliminary feature fusion. Specifically, the concatenated features are processed by an MLP layer and finally projected back to the channel dimension of the original base frame, using it as the new base frame. Then the base frame is transformed through a learnable linear layer to generate \(\Delta_{\text{base}_t}\) and \(B_{\text{base}_t}\), and these are then used in the discretization formula from Eq.~\eqref{eq:2} to obtain \(\overline{A}_{\text{base}_t}\) and \(\overline{B}_{\text{base}_t}\). The current frames are processed through another linear layer to obtain \(x_{\text{cur}_t}\), where \(t\) indicates the positional relationship after flattening the base frame and current frames.

Utilizing Eq.~\eqref{eq:2}, the zero-order hold and discretization can be expanded as follows:
\begin{equation}
\resizebox{0.9\linewidth}{!}{$
\begin{aligned}
h_t &= \sum_{j=0}^{t} \left[ \prod_{i=j+1}^{t} \exp(\Delta_{\text{base}_i} A) \right] \tilde{f}(\Delta_{\text{base}_j})B_{\text{base}_j}x_{\text{cur}_j}, \end{aligned}$}
\label{eq:5}
\end{equation}
\begin{equation}
\resizebox{0.9\linewidth}{!}{$
\begin{aligned}
y_t &= \scriptsize{C \sum_{j=0}^{t} \left[ \prod_{i=j+1}^{t} \exp(\Delta_{\text{base}_i} A) \right] \tilde{f}(\Delta_{\text{base}_j})B_{\text{base}_j} x_{\text{cur}_j}} \\
&\quad + \scriptsize{D x_{\text{cur}_t}},
\end{aligned}$}
\label{eq:6}
\end{equation}
where $\tilde{f}$ represents the functions of $\Delta$ and $B$ corresponding to the zero-order hold of $B$, as follows:
\begin{equation}
\resizebox{0.9\linewidth}{!}{$
\begin{aligned}
 \tilde{f}(\Delta_{\text{base}_t}) &=
 (\Delta_{\text{base}_t} A)^{-1} \left( \exp(\Delta_{\text{base}_t} A) - I \right)
 \Delta_{\text{base}_t}.
\end{aligned}$}
\label{eq:7}
\end{equation}
Specifically, in the State Space Model, the input \(x_t\) at time \(t\) is influenced by the control matrix \(B\), which in turn affects the change in state \(h\). In the discretized state space, the discretization step size \(\Delta\) represents the time \(x_t\) acts on the state. In QSSM, we desire the base frame to act as a gate controlling the influence of the current frames on the state, thereby affecting the output. Thus, we generate \(\text{base}_t\) and \(\Delta\) by the base frame through a linear layer. To exploit the differences in sub-pixels and noise distribution characteristics across multiple frames, we merge current features into the channel dim. Subsequently, all current features are projected into a unified set of merged current features through a linear layer. This allows the base feature to query all current features at once, thereby achieving multi-frame joint denoising. Additionally, as Eq. (\ref{eq:5}) and (\ref{eq:6}), when the flattened base feature \( f_{\text{base}_t} \) queries the current features at other times\( f_{\text{cur}_j} \), \( f_{\text{base}_t} \) modulates and guides \( f_{\text{base}_j} \) to query \( f_{\text{cur}_j} \) through the forget gate and input gate \(\exp(\Delta_{\text{base}_t} A)\), ultimately feeding back to the output at time \( t \). This enhances the interaction between \( f_{\text{base}_t} \) and its neighboring base features as well as current features. Due to the characteristics of the matrix \( A \), the influence of \( f_{\text{base}_t} \) in guiding the query of \( f_{\text{base}_j} \) gradually decreases with their distance in the sequence, forming a progressively diminishing receptive field. This prevents \( f_{\text{base}_t} \) from overly focusing on spatially distant information, thereby enhancing neighborhood interactions. Since each query by the base feature simultaneously queries all current features, the base feature can better perceive sub-pixel information consistently distributed across frames, suppressing random noise. We modify the RSSB block~\cite{c:8} by integrating our proposed QSSM with four scanning directions and using channel attention to enhance channel interaction. 

\begin{figure}[!t]
	\centering
\includegraphics[width=1.0\linewidth]{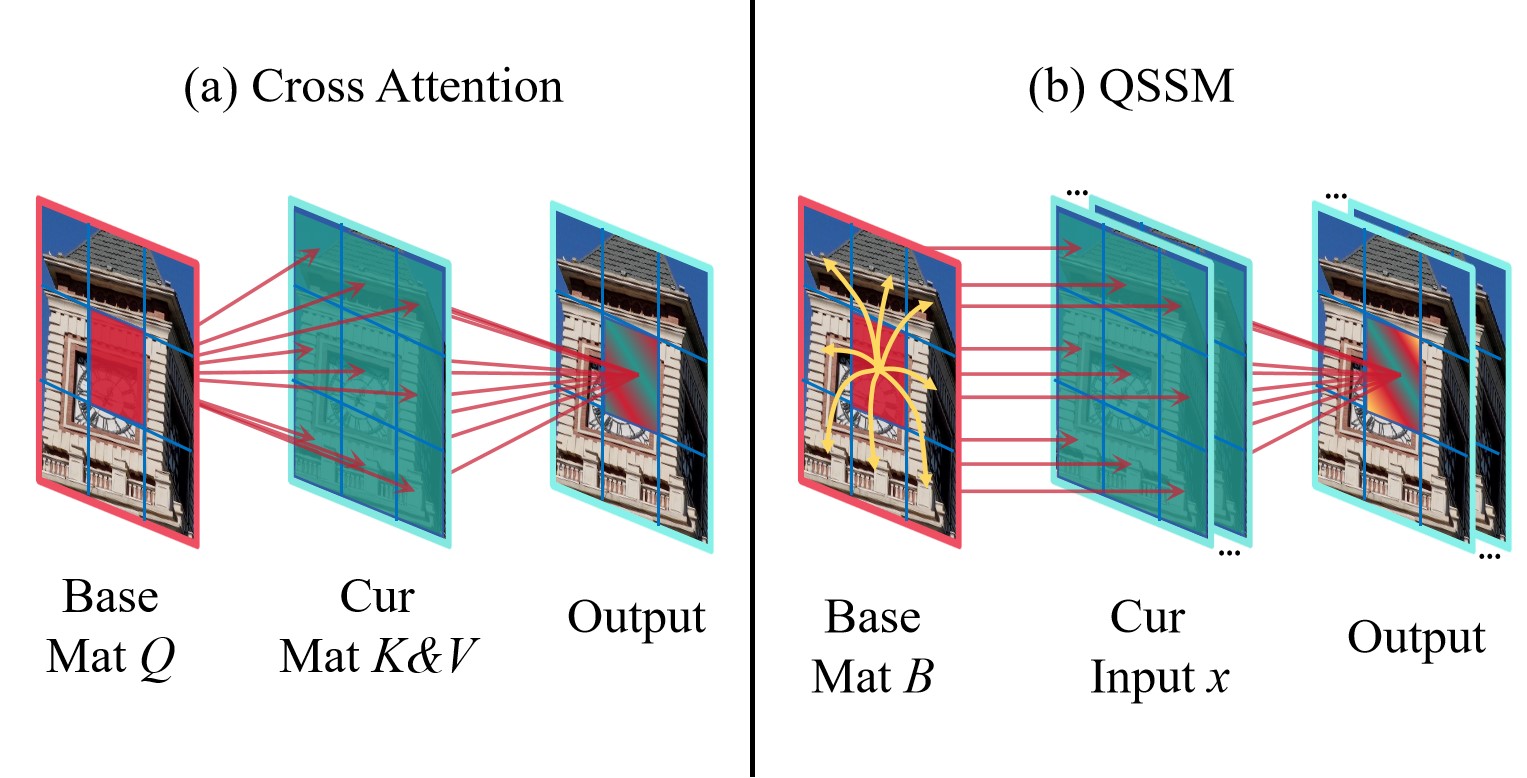}
\caption{Illustrative comparison between the proposed QSSM and existing cross-attention mechanisms.}		\label{fig:fig3}
\end{figure}
\noindent \textbf{Discussion} To clarify the differences between the proposed QSSM and existing cross-attention methods, we conduct a detailed analysis along with a schematic comparison. As shown in Figure~\ref{fig:fig3} (a), the cross-attention method uses serialized base feature tokens at position \( t \) as the Q matrix, querying with all position tokens of the current features as the K matrix to extract sub-pixel information. In contrast, our QSSM, as shown in Figure~\ref{fig:fig3} (b), not only extracts sub-pixel information from the current features to the base feature but also facilitates information interaction within the base feature. It simultaneously retrieves corresponding current feature tokens from other position tokens in the serialized base feature, thereby achieving joint multi-position sub-pixel information retrieval and noise suppression between the base feature and current features.

\subsection{Multi-Scale Fusion Module}

Considering the presence of sub-pixel information across various scales within the intricate details of images, we propose a novel Multi-scale Fusion Module (MSFM). This module is designed to effectively integrate multi-scale sub-pixel information from the current frames, thereby enhancing the capability for detailed image reconstruction. The MSFM comprises three distinct branches: a Convolutional Neural Network (CNN), a State Space Model (SSM) with diverse scanning orientations, and a channel Transformer. To begin with, a $3 \times 3$ convolution is utilized for the fusion of local sub-pixel features. The SSM is introduced to efficiently learn and integrate sub-pixel features along both horizontal and vertical axes. Furthermore, considering the attenuation characteristics of the A matrix within the SSM when dealing with long-range perception, we concurrently employ a Transformer block to augment the network's proficiency in capturing global information. The mathematical formulation of the MSFM is as follows:
\begin{equation}
y = w_1 \cdot \text{CNN}(x) + w_2 \cdot \text{SSM}(x) + w_3 \cdot \text{Transformer}(x)
\label{eq:8}
\end{equation} 
where \( w_{i} \) represents the balancing factors. \( x\) and  \( y\) represent the input feature and output feature, respectively.

\subsection{Adaptive Up-sampling Module}

{After the aforementioned processes, the sub-pixel structural information from burst LR images is extracted and distributed in the feature space. The next critical challenge is to utilize this valuable sub-pixel information to adaptively upsample the image resolution and incorporate sub-pixel details into the high-resolution image. Existing state-of-the-art methods, such as Burstormer~\cite{dudhane2023burstormer} and BIPNet~\cite{c:6}, simply employ interpolation, transposed convolution, or pixel shuffle techniques for resolution upsampling. However, these methods lack the capability to perceive the distribution of sub-pixels in the feature space, leading to an inability to adaptively reconstruct fine details. Therefore, we introduce a novel Adaptive Up-sampling (AdaUp) module that perceives the distribution of sub-pixels in the spatial domain and adaptively adjusts the up-sampling kernel, thereby achieving higher-quality image detail, as shown in Figure~\ref{fig:framework} (b).} Specifically, we first adaptively perceive the distribution of sub-pixels \( L\in\mathbb{R}^{B \times C_{\text{in}} \times 1 \times 1} \) from the input features \( X\in\mathbb{R}^{B \times C_{\text{in}} \times H \times W} \) by adaptive pooling. We then perform sequence feature interaction on \( L \) to obtain the output channel feature distribution sequence \( L_1\in\mathbb{R}^{B \times C_{\text{out}} \times 1 \times 1} \). Subsequently, we apply both the input distribution sequence and the output distribution sequence to the upsampling transposed convolution kernel \( W \in \mathbb{R}^{B \times C_{\text{in}} \times C_{\text{out}} \times 3 \times 3} \) using broadcasting, thereby endowing the kernel with feature perception capability. Finally, we obtain the high-resolution output through the upsampling transposed convolution kernel. The corresponding formulas are as follows:
\begin{equation}
L = \text{AdaptivePooling}(X)
\label{eq:9}
\tag{9}
\end{equation}
\begin{equation}
L_1 = \text{Conv}_{1 \times 1}(L)
\label{eq:10}
\tag{10}
\end{equation}
\begin{equation}
W_f = (W \odot L) \odot L_1
\label{eq:11}
\tag{11}
\end{equation}
\begin{equation}
y = \text{Trans-Conv}(W_f, X)
\label{eq:12}
\tag{12}
\end{equation}
where $\odot$ represents element-wise multiplication. Thus, AdaUp can leverage the underlying content information from input frames at different channels and utilize it to get better performance than the mainstream up-sampling operations, pixel shuffle, or interpolations~\cite{carlson1985monotone,schaefer2006image}. 

\section{Experiments and Analysis}

\begin{table*}[ht]
    \centering
    \setlength{\tabcolsep}{1.4mm}{
    \resizebox{1.0\linewidth}{!}{
    \begin{tabular}{lcccccccccccc}
        \toprule
        & {Bicubic} & {HighRes-net} & {DBSR} & {LKR} & {MFIR} & {BIPNet} & {AFCNet} & {FBAnet} & {GMTNet} & {RBSR} & {Burstormer} & \textbf{Ours} \\
        \midrule
        {PSNR}{$\uparrow$} & 36.17 & 37.45 & 40.76 & 41.45 & 41.56 & 41.93 & 42.21 & 42.23 & 42.36 & 42.44 & 42.83 & \textbf{43.12} \\
        {SSIM}{$\uparrow$} & 0.91 & 0.92 & 0.96 & 0.95 & 0.96 & 0.96 & 0.96 & 0.97 & 0.96 & 0.97 & 0.97 & \textbf{0.97} \\
        \bottomrule
    \end{tabular}}
    \caption{{Performance comparison of existing methods on Synthetic BurstSR dataset for ×4 BurstSR.}}
    \label{tab:comparison}}
\end{table*}

\begin{table}[t]
    \centering
    \setlength{\tabcolsep}{0.8mm}{
    \begin{tabular}{lcccccc}
        \toprule
        \multirow{2}{*}{{Method}} & \multicolumn{3}{c}{{RealBSR-RAW}} & \multicolumn{2}{c}{{RealBSR-RGB}} \\
        \cmidrule(lr){2-4} \cmidrule(lr){5-7}
        & {PSNR}{$\uparrow$} & {SSIM}{$\uparrow$} & {L-PSNR}{$\uparrow$} & {PSNR}{$\uparrow$} & {SSIM}{$\uparrow$} \\
        \midrule
        DBSR & 20.906 & 0.635 & 30.484 & 30.715 & 0.899 \\
        MFIR & 21.562 & 0.638 & 30.979 & 30.895 & 0.899 \\
        BSRT & 22.579 & 0.622 & 30.829 & 30.782 & 0.900 \\
        BIPNet & 22.896 & 0.641 & 31.311 & 30.655 & 0.892 \\
        FBANet & 23.423 & 0.677 & 32.256 & 31.012 & 0.898 \\
        Burstormer & 27.290 & 0.816 & 32.533 & 31.197 & 0.907 \\
        \midrule
        \textbf{Ours} & \textbf{27.558} & \textbf{0.820} & \textbf{32.791} & \textbf{31.401} & \textbf{0.908} \\
        \bottomrule
    \end{tabular}
    }
    \caption{{Performance comparison of existing methods on RealBSR-RGB and RealBSR-RAW datasets for ×4 BurstSR.}}
    \label{tab:combined_comparison}
\end{table}

\subsection{Experimental Settings}
\begin{figure*}[ht]
	\centering
	\includegraphics[width=1.0\linewidth]{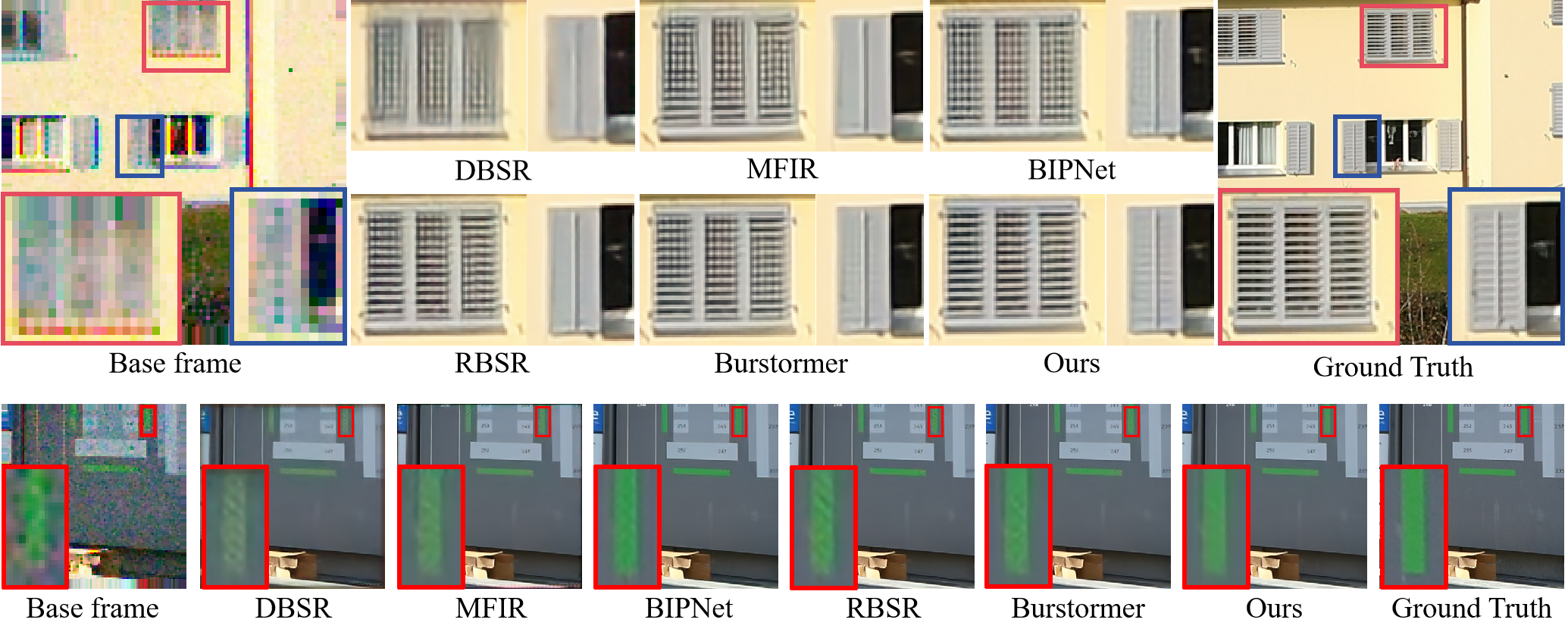}
	\caption{Visual comparison results with different methods on SyntheticBurst datasets for ×4 BurstSR.}
	\label{fig:syn}
\end{figure*}

\begin{figure*}[ht]

	\centering
	\includegraphics[width=1\linewidth]{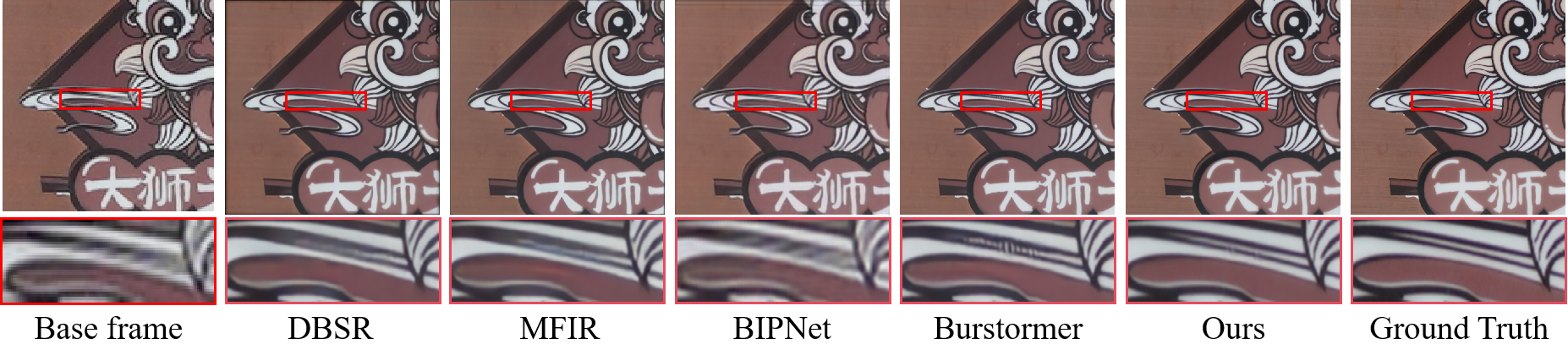}
        \vspace{-4mm}
	\caption{Visual comparison results with different methods on RealBSR-RGB dataset for ×4 BurstSR.}
     \vspace{-1mm}
	\label{fig:srgb}
\end{figure*}

\textbf{Implementation details.} 
{We evaluate the effectiveness of our proposed method on four public burst image super-resolution benchmarks, encompassing both synthetic and real datasets: synthetic BurstSR~\cite{bhat2021deep}, Real BurstSR~\cite{bhat2021deep}, RealBSR-RAW~\cite{wei2023towards}, and RealBSR-RGB ~\cite{wei2023towards}. To ensure fairness, we follow~\cite{dudhane2023burstormer} for training and evaluation. More details of datasets and data processing can be found in Appendix A section. Following~\cite{dudhane2023burstormer}, we train the model from scratch on the synthetic BurstSR dataset for 300 epochs, using the AdamW optimizer with parameters $\beta_1 = 0.9$ and $\beta_2 = 0.999$. We employ a cosine annealing strategy to gradually decrease the learning rate from $3 \times 10^{-4}$ to $10^{-6}$ and set the training patch size to $48 \times 48$. For the Real BurstSR dataset, we follow~\cite{bhat2021deep} to fine-tune the model pre-trained on synthetic BurstSR for 60 epochs, maintaining the same training setting as the synthetic BurstSR but adjusting the learning rate to $1 \times 10^{-6}$ and the training patch size to $56 \times 56$. For the RealBSR-RAW and RealBSR-RGB datasets, we follow~\cite{wei2023towards} to train from scratch for 100 epochs, using the same training setting as the synthetic BurstSR, with a training patch size of $80 \times 80$. We set the batch size to 8, and the burst size to 14, and all experiments are conducted on 8 V100 GPUs.}

\noindent \textbf{Metric.}  
{Following previous works~\cite{bhat2021deep,wei2023towards}, we use reference metrics to evaluate performance, including PSNR and SSIM.}

\noindent \textbf{Compared methods.}
To comprehensively demonstrate the superiority of our proposed method, we compare our QMambaBSR with ten classic and state-of-the-art (SOTA) BurstSR approaches HighRes-net~\cite{c:10}, DBSR~\cite{bhat2021deep}, LKR~\cite{c:11}, MFIR~\cite{a:23}, BIPNet~\cite{c:6}, AFCNet~\cite{c:12}, FBAnet~\cite{wei2023towards}, Burstormer~\cite{dudhane2023burstormer}, RBSR~\cite{wu2023rbsr}, GMTNet~\cite{c:7}.

\subsection{Quantitative and Qualitative Results}
{\noindent\textbf{Results on the Synthetic BurstSR dataset.} As shown in Table~\ref{tab:comparison}, our method outperforms existing BurstSR methods, achieving the best performance. For example, compared to the existing SOTA method, Burstormer, our method achieves a 0.29 dB improvement in PSNR. Furthermore, to further demonstrate the visual superiority of our method, we present a visual comparison with existing methods in Figure~\ref{fig:syn}. We can observe that the RAW low-resolution images exhibit significant noise and severe detail loss, as illustrated in the window area of Figure~\ref{fig:syn}. Compared to existing methods, our method demonstrates superior performance in reconstructing textures and details in the window area. Additionally, in the green stripes region at the bottom of Figure~\ref{fig:syn}, the substantial noise in the base frame leads existing methods to leave artifacts. However, our method more effectively distinguishes between noise and sub-pixels, producing more detailed, artifact-free high-resolution images, thereby demonstrating the visual superiority of our method.
}

{\noindent\textbf{Results on RealBSR-RGB and RealBSR-RAW.} As shown in Table~\ref{tab:combined_comparison}, our method consistently outperforms existing methods on these two real benchmarks, achieving the best performance. For RealBSR-RAW, our method surpasses FBANet and Burstormer in PSNR and linear-PSNR by 0.268 dB and 0.258 dB, respectively. For RealBSR-RGB, our method surpasses FBANet and Burstormer in PSNR by 0.204 dB. Furthermore, as shown in Figure~\ref{fig:srgb}, the results demonstrate the superior performance of our method in detail reconstruction and artifact suppression. This validates the effectiveness of our method in real-world scenarios, highlighting its superiority and practicality. More results on Real BurstSR and qualitative results will be presented in the appendix.
}

\subsection{Ablation Study}

{To demonstrate the effectiveness and superiority of the proposed modules, we conduct a series of ablation experiments. Specifically, we incrementally integrate the proposed modules into the baseline network. For rapid evaluation, we train our model on the synthetic dataset for 100 epochs. From Table~\ref{tab:performance_comparison}. we observe that the introduction of the MSFM module, which enhances the network's multi-scale perception and fully integrates sub-pixel information from different frames, significantly improves performance by 1.34 dB in PSNR. Furthermore, the addition of the QSSM, which extracts sub-pixels from the current frames that match the content of the base frame while suppressing noise, leads to an additional performance gain of 0.72 dB in PSNR. Finally, incorporating the proposed Adaptive Up-sampling module, which better adapts the up-sampling kernel according to the scene, thereby generating high-resolution images with richer details, results in a further improvement of 0.26 dB. These results indicate that the proposed modules significantly enhance burst super-resolution performance.}

{\noindent\textbf{Comparison with Existing Modules.} To verify the effectiveness of proposed module, we replaced it with existing fusion and up-sampling modules. To effectively represent the performance of various methods while also considering training time, we train for 100 epochs on the synthetic dataset. As shown in Table~\ref{tab:exp2}, in the fusion stage, compared to PBFF~\cite{c:6} or NRFE~\cite{dudhane2023burstormer}, our QSSM and MSFM modules are able to better exploit the inter-frame consistency of sub-pixel distribution while denoising, resulting in PSNR improvements of 1.56 dB and 0.41 dB, respectively. In contrast to static upsampling like pixel shuffle and transposed convolution, our AdaUp module enhances the network's ability to perceive scene-specific sub-pixel distributions, leading to a PSNR improvement of 0.16 dB.
}
\begin{table}[t]
    \centering
    {\fontsize{11.5}{12}\selectfont 
    \begin{tabular}{ccccc}
        \toprule
        \textbf{QSSM} & \textbf{MSFM} & \textbf{AdaUp} & \textbf{PSNR}{$\uparrow$} & \textbf{SSIM}{$\uparrow$} \\
        \midrule
        × & × & × & 39.81 & 0.93 \\
        × & \checkmark & × & 41.15 & 0.94 \\
        \checkmark & \checkmark & × & 41.87 & 0.96 \\
        \checkmark & \checkmark & \checkmark & \textbf{42.13} & \textbf{0.96} \\
        \bottomrule
    \end{tabular}
    }
    \caption{Ablation experiment on proposed core modules.}
    \label{tab:performance_comparison}
\end{table}

{\noindent \textbf{Evaluation about MSFM.}} To validate the effectiveness of different branches within the proposed MSFM, we conducted ablation experiments with various internal branch setting. The experimental results are shown in Table~\ref{tab:evaluation_msfm}. To minimize training time while ensuring experimental validity, the ablation experiments were conducted on a synthetic dataset and trained for 100 epochs. As shown in the table, we use only the convolution in the MSFM module as the baseline. Adding the transformer module resulted in a PSNR increase of 0.39 dB. When the transformer was replaced with SSM, the PSNR improved by 0.44 dB compared to using only the conv method. Under our proposed MSFM, the PSNR increased by 0.56 dB, clearly demonstrating the effectiveness of both the SSM and transformer branches in the MSFM.

\begin{figure}[t]
	\centering
	\includegraphics[width=1\linewidth]{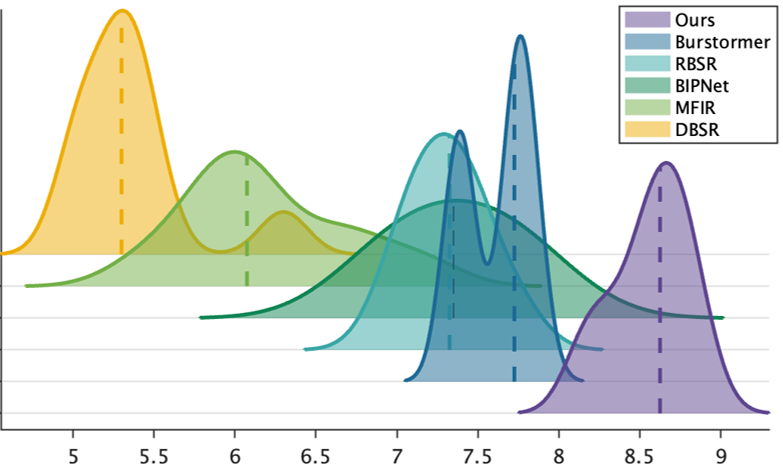}
	\caption{User study of reconstructed real HR images.}
	\label{fig:user_study}
\end{figure}

\begin{table}[t]
\centering
{ 
\begin{tabular}{l l c c}
\toprule
 \textbf{Process Stage}& \textbf{Methods} & \textbf{PSNR}{$\uparrow$} & \textbf{SSIM}{$\uparrow$} \\
\midrule
\multirow{3}{*}{Fusion} & Concat & 39.85 & 0.93\\
                        & PBFF~\cite{c:6} & 40.57 & 0.94\\
                        & NRFE~\cite{dudhane2023burstormer} & 41.72 & 0.96\\
\textbf{} & \textbf{Ours} & \textbf{42.13}  & \textbf{0.96} \\
\midrule
\multirow{3}{*}{Up-sampler} & Pixelshuffle & 41.89 & 0.96 \\
                            & Transposed conv & 41.97 & 0.96 \\

\textbf{} & \textbf{Ours} & \textbf{42.13} & \textbf{0.96} \\
\bottomrule
\end{tabular}
}
\caption{Comparison with existing modules on fusion and upsampling process stage.}
        \vspace{-4mm}

\label{tab:exp2}
\end{table}

\subsection{User study}

{To effectively illustrate the superiority of our proposed method in reconstructing visually pleasing images, we conducted a user study involving 10 real burst images selected from well-established benchmarks. Twenty volunteers participated in this study, tasked with rating the similarity and quality between each reconstructed image and the ground truth (GT). They used a detailed scale ranging from 0, indicating visually unsatisfactory and completely dissimilar images, to 10, representing visually satisfactory and highly similar images. We then meticulously aggregated the scores from all volunteers, and the results are depicted in Figure~\ref{fig:user_study}. When compared to existing methods such as Burstormer and BIPNet, our proposed method stands out by adaptively extracting sub-pixels to achieve superior visual effects. This approach earned our method the highest average score of 8.56, underscoring its effectiveness.}

\begin{table}[t]
    \centering
    {
    \begin{tabular}{ccccc}
        \toprule
        \textbf{Conv} & \textbf{Transformer} & \textbf{SSM} & \textbf{PSNR}{$\uparrow$} & \textbf{SSIM}{$\uparrow$}\\
        \midrule
        \checkmark & × & × & 41.57 & 0.95\\
        \checkmark & \checkmark & × & 41.96 & 0.96\\
        \checkmark & × & \checkmark  & 42.01 & 0.96\\
        \checkmark & \checkmark & \checkmark & \textbf{42.13} & \textbf{0.96} \\
        \bottomrule
    \end{tabular}
    }
    \caption{Ablation studies on our proposed MSFM.}
        \vspace{-4mm}
    \label{tab:evaluation_msfm}
\end{table}

\section{Conclusion}
In this paper, we introduce a novel approach called QMambaBSR for burst image super-resolution. Based on the structural consistency of sub-pixels and the inconsistency of random noise, we propose a novel Query State Space Model to efficiently query sub-pixel information embedded in current frames through an intra- and inter-frame multi-frame joint query approach while suppressing noise interference. We introduce a Multi-scale Fusion Module for information on sub-pixels across different scales. Additionally, a novel Adaptive Up-sampling module is proposed to perceive the spatial arrangement of sub-pixel information in various burst scenarios for adaptive up-sampling and detail reconstruction. Extensive experiments on four public synthetic and real benchmarks demonstrate that our method surpasses existing methods, achieving state-of-the-art performance while presenting the best visual quality.

In future work, we plan to utilize state space models further to enhance the alignment stage in burst image super-resolution and explore more efficient and coherent integration of our architecture across various vision tasks. Additionally, we aim to unlock our method to other burst image restoration tasks, such as denoising, HDR, and more.

{
    \small
    \bibliographystyle{ieeenat_fullname}
    \bibliography{main}
}
\end{document}